\documentclass[conference]{IEEEtran}
\usepackage{cite}
\usepackage{fixme}
\usepackage{enumitem}
\ifCLASSINFOpdf
  \usepackage[pdftex]{graphicx}
\else
  \usepackage[dvips]{graphicx}
\fi
\usepackage{amsmath}
\usepackage{url}

\begin{document}

\listoffixmes

\title{Automated Game Design Learning}
% \subtitle{}

\author{\IEEEauthorblockN{Joseph C.\ Osborn, Adam Summerville, and Michael Mateas}
\IEEEauthorblockA{Computational Media\\
University of California, Santa Cruz\\
Santa Cruz, CA 95064\\
Email: jcosborn@ucsc.edu}
%\and
%\IEEEauthorblockN{Christoffer Holmgard}
%\IEEEauthorblockA{...}
%\and
%\IEEEauthorblockN{Daniel Zhang}
%\IEEEauthorblockA{...}
}

\maketitle

\begin{abstract}
  While general game playing is an active field of research, the learning of game design has tended to be either a secondary goal of such research or it has been solely the domain of humans.
  We propose a field of research, Automated Game Design Learning (AGDL), with the direct purpose of learning game designs directly through interaction with games in the mode that most people experience games: via play.
  We detail existing work that touches the edges of this field, describe current successful projects in AGDL and the theoretical foundations that enable them, point to promising applications enabled by AGDL, and discuss next steps for this exciting area of study.
  The key moves of AGDL are to use game programs as the ultimate source of truth about their own design, and to make these design properties available to other systems and avenues of inquiry.
\end{abstract}

\section{Introduction}
\begin{figure}[hbtp!]
\centering
\includegraphics[width=0.48\textwidth]{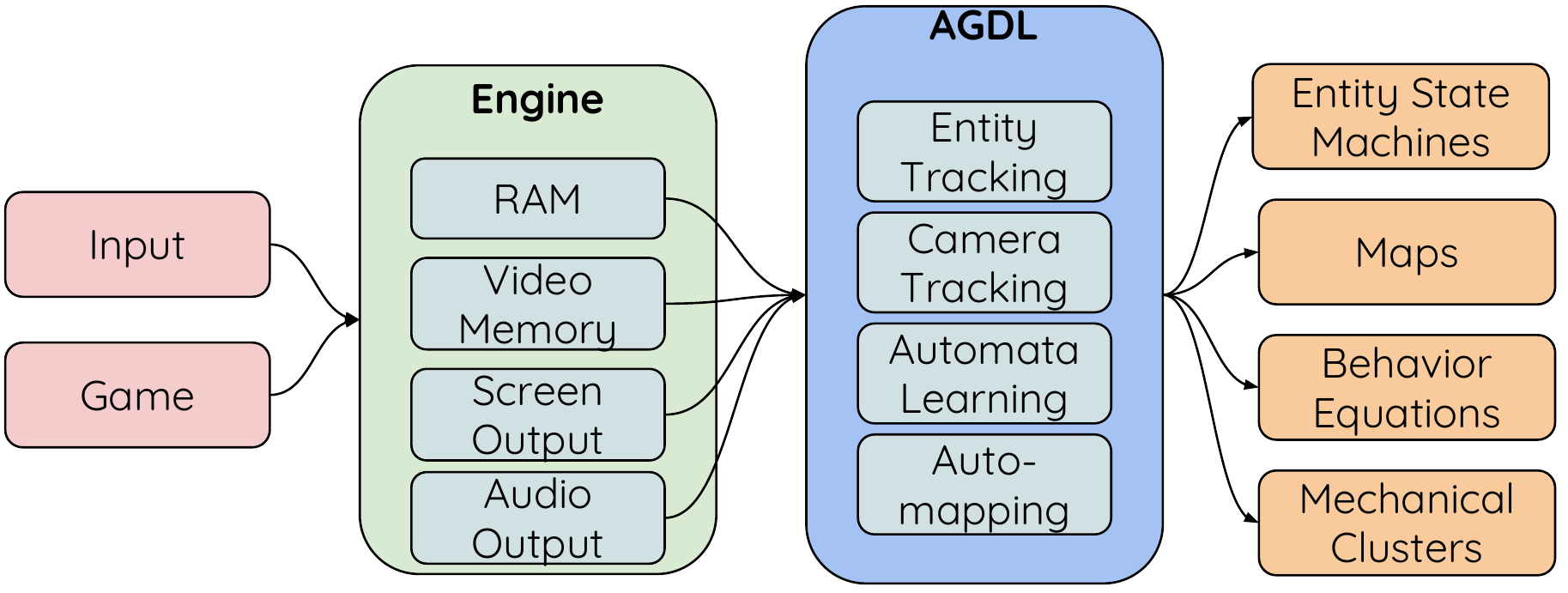}
\caption{Proposed pipeline for AGDL.  Player input can come from sources of opportunity (e.g.\ speed runs), AI players, or human playtesting. Different game engines will expose different sets of observations which feed into the learning process (some of which are listed).  Various learning modules (Blue sub-components) can be designed to produce relevant corresponding outputs (Orange).  This list is not exhaustive, but shows some of the features, techniques, and outputs we have successfully used in recent projects.}
\label{fig:AGDL_Schema}
\end{figure}
Games are full of complex and emergent behaviors, and their runtime behavior is hard to predict from their source code or binaries. The game code \emph{is} the design in the ultimate way, but neither the designer nor the player have direct access to this design.
In general, source code is specified at too fine a grain (frame-by-frame transition rules) and in a way which is hard for humans to work with.  Instead, game creators (programmers and designers) and players each have a model of the game with which the code may or may not agree.
Both groups learn more about the game as they play it.
Even an initial brush against a game reveals an abstract specification concerning high-level design elements: characters' appearances, animations, and general behaviors; level layout and connections; a game's goals; and so on.

In this paper, we distinguish the \emph{game program} (the low-level rules that are the ultimate source of truth on the game but are hard for machines and people to reason about) from the \emph{game design}, with the understanding that a design that is not at the level of source code is necessarily incomplete or highly abstracted (with many possible low-level implementations).
%Undeniably, a game maker explores the space of possible designs in the process of programming and building the game.
Our division between the design and the program is not meant to imply that the \emph{true} game is fully-formed in a designer's head before  \emph{translation} to code; rather, that game makers, being human and not machines, necessarily have a different and more abstracted understanding of their code (else software bugs and surprising emergent behavior would not exist).
Fortunately, many design properties of interest can be tested by looking only at the abstract design, regardless of the low-level implementation.

In this paper, we propose a new research agenda, Automated Game Design Learning (AGDL), rooted in several existing research traditions.
Advances in AGDL can directly benefit other areas of interest in games and AI; moreover, AGDL is in and of itself a productive field of study.
We envision several key application areas:

\begin{description}[align=left]
\item [Improving Human Play] --- E.g., seeing a fighting game character's state machine, collision frames, and so on is useful for players.
\item [Improving General Game Play] -- E.g., learning to construct useful abstractions to guide planning.
\item [Quantitative Game Studies] --- E.g., comparing mechanics across games or pulling out game level data for architectural study.
\item [Data-Driven Procedural Content Generation] --- E.g., learning the mechanical properties of the tiles and entities in a game for corpus generation purposes.
\item [Code Studies] --- E.g., tracing high level design rules down through runtime behavior to understand source (or machine) code.
\item [Game Design Verification] --- E.g., extracting formal models of game rules and ensuring that certain design properties hold on the model.
\end{description}

%\subsection{How could it be automated?}

How do we extract a game design from a game program?
Often, human designers do it by reading code and thinking, but reading code in any way besides just executing it is in general hard for computers (and many very different programs could be behaviorally identical).
We therefore focus on the way that most people learn a game's design: \emph{play}.
Players and designers (playtesting being a crucial activity) both  interact with and learn a game's rules through playing, experimenting, and thinking---this is the mechanism we propose to automate.

Our goal is to extract high level design properties like level structure, character behaviors, progression, resource exchanges, and so on from observations of human or AI play in as automatic a manner as possible.  This is an open-ended problem and, at the fully general level, it is surely undecidable; nonetheless, we discuss ways we have found to make the problem tractable in several interesting cases.

Our starting point is a knowledge representation founded in Operational Logics (OLs)~\cite{mateas_defining_2009}.
We are not treating games as bags of mechanics (e.g.\ we do not operate at the level of saying ``this game is a platformer'' or even ``this type of collision (always) triggers that type of reaction'').
Instead, we learn in terms of individual OLs' abstract operations---what mechanics are made from.
The communicative strategies provided by operational logics give us a clear way to connect what a game \emph{looks} like with how a game \emph{works}.
Sec.~\ref{sec:ol} explores this approach in detail.

For the works presented here, we target games on the Nintendo Entertainment System (NES) for several reasons:
\begin{itemize}
\item It hosts a diverse set of games
\item The architecture offers a clean way to identify high level image features easily
\item It has an active community providing a library of playthroughs of these games
\item Efficient software emulators offer fine-grained control and capture of audiovisual and other state
\end{itemize}

%We are not looking at reimplementations of games like \emph{Infinite Mario Bros.}; instead we interrogate the \emph{actual} \emph{Super Mario Bros.}\ as the source of truth about \emph{Super Mario Bros.}

% \subsection{Automated Game Design Learning}

In this paper, we review a broad range of related work both in games and in general automated software analysis, from communities as diverse as game enthusiasts, game scholars, and researchers in software testing and verification.
After surveying the fields from which the present work draws, we describe a foundation for representing game design knowledge in a genre-independent way.
We follow by briefly introducing several successful projects in learning game character behaviors and structural properties of game levels from NES games.
Finally, we show some possible directions for extending the present work's domain of discourse and applying its results more widely.

\section{Related Work}

\subsection{Game Design Support}

Game design support provides a great research challenge.
Beyond the simplest games, games are complex emergent systems where it is hard to predict the broad outcomes of even small rules changes.
There are many ``filters'' for this in the process of game design: 
The designers' intuitions are a first filter, designer playtesting a second, ``fresh player'' playtesting a third, and play community testing a fourth filter.
Each filter tests for different things~\cite{nelson_requirements_2009}, but some of these cases can be handled by playtesting without players~\cite{nelson_interactive_2008} and this approach can be generalized in principle~\cite{smith_open_2013}.

Oftentimes, direct search on the low level transition systems of games is intractable.
We therefore often analyze abstractions of games that we produce manually.
For example, we call the special case of navigating a character through a virtual space ``pathfinding''.
We often implement pathfinding by abstracting the game's movement rules onto a gridded or other schematic space, doing search on that higher level representation, and then grounding out the results into low-level action plans. % (sometimes these plans are not realizable, and sometimes there are legitimate paths which are not found in these approximations).
This encodes designer knowledge into explicit abstractions, in a sense ignoring the behavior of the actual program.
The automatic calculation of reachable regions of a game space from designers' knowledge of physics rules is at the core of navmesh-based approaches to pathfinding~\cite{arkin1987path}.
At the same time, convex decompositions often abstract the physics rules too much, so some have used under-approximations like random search to find what parts of the possibly-reachable space the game's true dynamics can actually reach~\cite{walkmonster,tremblay2013exploration}. 

Many successful projects in game analysis have used these manual abstractions.
Bauer's work on Treefrog Treasure~\cite{bauer2012rrt} moved from directly driving the game code towards path planning via parabola-line segment intersections (in fact, this abstraction later made it back into the game code).
In this case, treating the game character as a point was an over-approximation that simplified search without introducing too many false paths.
Shaker \textit{et al.}'s editor for Cut the Rope levels used polyhedral over-approximations to show which parts of the stage were influenced by particular puzzle elements (essentially finding closed-form solutions to aspects of the puzzle)~\cite{shaker2013automatic}.
Smith \textit{et al.}'s tools for the Refraction educational game~\cite{smith2012case,smith_quantifying_2013} use a tight over-abstraction of the game's core rules where the order in which puzzle pieces are placed does not matter, whereas the qualitative spatial relations between those remains important; this is key to keeping the encodings of puzzles and solutions small.

Creating these abstractions is important, but it is ad hoc, game-specific, and labor-intensive.
Essentially, they are redundantly describing the design in several places: once in the code, and once for each abstraction over which the tools operate.
In theory a high level design could be written and then refined in a principled way, but designers tend to work in more exploratory ways, responding to prompts like unexpected bugs or the implicit definitions of partially-implemented programs.
One motivation for AGDL is finding these useful abstractions automatically, so that designers don't have to learn modeling languages or take the time to define several versions of their game design and keep them all in sync.
Moreover, if a designer sees an abstraction which does not agree with their own understanding of the game's rules, then they have discovered a bug.

\subsection{G(V)GP}
\label{sec:ggp}

Automated game design learning is explicitly \emph{not} General Game Playing (GGP).
We are not trying to achieve optimal play, and only care about rewards to the extent that they reveal truths about the design.
Of course, we borrow from GGP and General Video Game-AI (GVG-AI), and insofar as learning the design of the game can help produce a more intelligent general player the contributions of AGDL can feed into further GGP and GVG-AI research.

The two main areas of inquiry which GGP has begun to explore and which AGDL generalizes and could support more broadly are heuristic learning and transfer learning.
Generally speaking, heuristic learning is a way to learn, on a game-by-game basis, about intermediate goals of the game or rough strategies for guiding search.
The GGP agent FluxPlayer made its name by statically analyzing game rules to automatically determine a position evaluation function to be used as a heuristic~\cite{schiffel2007fluxplayer}; i.e., from just the rules it could make an estimate as to how close or far from the goal a particular game state was.
FluxPlayer also established which aspects of the rules defined high-level structures like ordered relations, quantities of game resources, board positions, and so on, feeding those into the distance function used for heuristic calculation.

In general videogame playing too, heuristic methods have become quite successful~\cite{perez20162014,perez2016general}.
Some agents match heuristics from a fixed portfolio against the game they are playing, and several agents try to determine whether each non-player character in the game is dangerous or desirable.
These can be seen as essentially both GVG-AI and AGDL agents, in that they are trying to learn about the game's design (AGDL) in order to play it effectively (GVG-AI).

Apart from learning heuristics for a single game, transfer learning is a key area where the portable design representations learned by AGDL could be of use to GGP agents.
Transfer is a key aspect of human learning, and indeed human game playing relies heavily on literacies obtained by playing other games in the past.
Banerjee and Stone extracted high-level features from the value functions learned on one GGP game to accelerate learning in another~\cite{ICML06-bikram} . 
Outside of the GGP context, 
%%Konidaris et al ????
K\"onik \textit{et al.}\ learn ``tasks'' suitable for transfer within the current environment (if the same problem comes up again later on) or to other environments with similar tasks by biasing value functions when  similarities are found~\cite{konik_skill_2009}.
We believe AGDL could  support this latter type of learning, giving a way to bridge high-level design concepts across games and to learn tactics and high-level actions within a given game.
One promising approach due to Braylan and Miikkulainen~\cite{braylan2016object} is to leverage design concepts like \emph{game characters} to  modularize reinforcement learning models, in the process admitting transfer learning by analogizing characters across games.

Another interesting thread is the general play of essentially opaque commercial games, as in for example the Arcade Learning Environment~\cite{bellemare2015arcade}.
While learning directly from visual features is important research, here we focus on a particular pair of projects: Learnfun and Playfun~\cite{murphy2013learnfun}.
Learnfun observes human play of a given NES game to learn both macro-actions representative of realistic inputs for that game \emph{and} a lexical order over RAM locations, with the intuition that success in many games can be characterized by a number or a set of numbers increasing.
For example, progress in \emph{Super Mario Bros.}\ is measured (roughly) by observing which world and level the player is in, and then by horizontal position within the level.
Its counterpart \emph{Playfun} uses those learned macro-actions in a heuristic search process attempting to optimize that lexical order.
Of course there are games for which this process works very poorly, and it will tend to react to an impending loss by pausing the game and halting, but it is surprisingly effective at surfacing useful features.

\subsection{Manual reverse-engineering techniques}

Reverse-engineering design properties from games is challenging even for humans, but people are highly motivated to do this.
Normally, a user has nothing more than a video of play---or, if they're lucky, a ROM---and want to get at games' audiovisual assets or other features.
This can be motivated by a desire to remix the game's designed elements in various ways; to learn subtle aspects of the game's rules; to cheat at the game; or to modify the game itself, e.g.\ to translate it into another language or change its levels, characters, or even rules.

The most popular activities around the extraction of game design elements from game programs are likely the manual processes of \emph{ripping} sprites, tilesets, 3D models, and other audiovisual assets from games' code and included resources.
Using screenshots and image editing programs, enthusiasts laboriously copy and paste game characters' animation frames into composite images~\cite{spriters_resource} or stitch together  screenshots into full level maps~\cite{vgatlas}.
This latter activity has been attacked with greater rigor in the form of the Video Game Level Corpus~\cite{summerville2016vglc}, which establishes standardized level formats.

Game players, especially competitive players, are also interested in learning about a game's rules.
Textual walkthroughs can be seen as a high-level summary of a game design, and the resources accumulated by fighting game enthusiasts~\cite{hitboxviewer,onesmash} are essentially reverse-engineered specifications of the game design suitable for study and practice.

The community of game \emph{speed-runners} have a special interest in the connection between the high-level game design and the low-level rules that enact it.
Here, the question of interest is essentially a classical combinatorial optimization problem: is there a sequence of button inputs of length \(L\) such that the objective \(z\) (for example, the length of \(L\) or the number of distinct button presses) is minimized?
A sub-community of \emph{tool-assisted} speed runners explicitly write such sequences of inputs one by one in specialized text editors, and both need and develop extremely deep knowledge about games' true, source code-level design features.
Recently, a YouTube video describing how details of modulo arithmetic enable the completion of a \emph{Super Mario 64} level with ``\(0.5\) A-button presses'' obtained millions of views~\cite{rollingrocks}.
The accompanying image gallery shows the process of discovering this sequence of moves and the tools and techniques built to support it~\cite{rollingrocksIMG}.

The FCEUX emulator for Famicom and NES has a variety of tools to support this community, including memory comparison interfaces, debuggers, and Lua integration to drive the emulator's main loop for experimentation~\cite{fceux}.
Enthusiasts of individual games also produce purpose-built tools for viewing and modifying those games' data structures in design-relevant ways; for example, to change their levels, their characters' statistics and appearances, or even plot events and script sequencing~\cite{romhacking}.
This all depends on knowing how these design elements are realized in the underlying game code.

\subsection{Automated game reverse engineering}

Because the manual processes outlined above are so game-specific on top of being tedious and labor-intensive, there have been some efforts to directly analyze games automatically to extract design-relevant information.
Martens \textit{et al.}~\cite{martens2016proceduralist} used Answer Set Programming (ASP) to perform ``proceduralist readings''  ~\cite{treanor_proceduralist_2011} automatically.
Given a specification of a game, their system applies a series of logic rules to deduce the readings that might be present in a game.
Using these ``meaning derivations'', they successfully build up low-level observations about a game (the player controls sprite $A$) into mid-level inferences (the player will attempt to make sprite $A$ collide with sprite $B$ because that will increase the resource that prevents them from losing) all the way up to to high-level interpretations (the game is futile since the difficulty increases monotonically).
This can be seen as an alternative application of the approaches to automatically finding heuristic functions outlined in Sec.~\ref{sec:ggp}.

There are also techniques that work directly on black-box game programs which inspired the AGDL project.
Murphy's \emph{glEnd() of Zelda} project aimed to automatically present a first-person 3D view of 2D NES games, without modifying the games~\cite{murphy2016glend}.
This requires knowledge of a game's design: whether the camera is top-down or side-view, whether gravity exists, which sprite on the screen is the player, and so on.

\emph{glEnd()} explored two especially interesting ideas: First, to look at the NES's graphics processor's memory rather than the output pixels to get certain high level features for free;
and second, to perform guided experiments that prove properties of interest and find parameters of the game's design.
For example, to determine which sprite is controlled by the player we might compare how much the position of each sprite on screen changes after holding left, holding right, or standing still from the same start state.
To tie locations in system RAM to sprites' on-screen positions, we could find all locations which have the same integer value as that position and see if modifying each such address once and then waiting a frame causes the sprite to draw in a different place (here, the intuition is that one RAM location determines the others).
We could determine whether gravity applies to a character by placing it in the sky (by modifying RAM) and then resuming play without providing any other inputs; if the character falls, then gravity applies to it.
%We could figure out whether a tile blocks movement by putting the player next to it and attempting to move the character.
%Apart from the existence proof that recovering game design properties from game binaries is possible, \emph{glEnd() of Zelda} provides an important insight: even if some game situations are impossible (e.g., placing the character in an unreachable part of the screen), they might still be \emph{informative}.
Murphy's work is effective for the automatic 3D-ification of NES games, but how can we generalize it to other kinds of properties or make it more robust?

Our earlier work~\cite{summerville2017what} attempted to learn ``latent causal affordances'' of game entities via observation of collisions and the potential effects of those collisions.
We applied a variety of machine learning techniques to cluster entities based on their observed mechanical properties, learning things like ``these objects all occlude Mario's motion'' which we would summarize as ``solid''.

Guzdial and Riedl learned implicit rules of \emph{Super Mario Bros.}\ level design by observing videos of gameplay~\cite{guzdial_toward_2015}.
Machine learning techniques were used to build probabilistic models which captured level design knowledge like \emph{treetop tiles should be above a rectangle of tree trunk tiles}.
With the same video source as input, Summerville \textit{et al.}~\cite{summerville2016learning} leveraged player path information to bias the generation of a level towards the specific play style of a given player.
Using recurrent neural networks they learned level design and player paths simultaneously, allowing the generator to bias implicitly the generation towards actions it saw (e.g.\ if a player interacted with question mark blocks it generated more of those blocks to make the generated path more likely to interact with them).

In the general game playing domain, techniques for learning game rules from observations of game behavior have been explored in recent years.
Bj\"{o}rnsson's Simple Game Rule Learner~\cite{bjornsson2012learning} derives the rules for a restricted class of Game Description Language (GDL) games by learning finite-automaton models for each of the game's pieces.
Gregory \emph{et al.}\ extend this work by incorporating techniques from the planning domain model acquisition literature~\cite{gregory2015grl}, learning interactions between pieces and admitting the dynamic addition and removal of pieces.

%% I'm not really sure that there's a strong case for much else in this section?
%%\fxnote{Other data-driven PCG stuff, Adam?}

\subsection{Specification Recovery From Software}

The problem of automatically extracting a specification from a software system is of general interest.
Variously called specification mining~\cite{dallmeier2010generating}, design recovery~\cite{biggerstaff1989design}, or just reverse engineering~\cite{chikofsky1990reverse}, the general schema is the same: looking at a program's source code or runtime behavior, synthesize an abstracted model of the software suitable for analysis.

Traditional approaches to specification recovery of opaque software include automata-based methods~\cite{shoham2008static} and the use of inductive logic or statistical learning~\cite{cohen1995inductive}.
If source code is available, pattern-matching techniques over the code can be used to obtain high-level design information~\cite{shi2006reverse} or even connect source code and corresponding documentation~\cite{marcus2005recovery}.
Abstract interpretation (including symbolic execution) and instrumented fuzzing are two key components of modern dynamic analysis systems, enabling search over the space of possible program paths; tools like SAGE~\cite{sage} and angr~\cite{shoshitaishvili2016state} effectively combine these to deeply explore programs' runtime behavior.
Each system targets particular sets of design properties and necessarily either over- or under-approximates systems' true behaviors, missing some aspects of the design or imputing others where they are not truly present.

\section{Knowledge Representation} 

Specification recovery is essentially unsolvable due to the halting problem and to the unbounded space of possible specifications that a given program refines.
Why do we suppose that we can make progress in this area in the domain of games?
Games differ from arbitrary software in two key ways.
First, games must be legible to players (and indeed their own designers): it must be possible for players to form reasonably accurate models of a game's behavior, or the game is unplayable.
Second, there are diverse and mature theories of game design to inspire and organize knowledge representation.

Researchers have long known that picking the correct problem specification and knowledge representation is vital for a system's success~\cite{agre_computation_1997}.
A successful game design learning system (as distinct from a generic specification recovery tool) must be parameterized by a way of organizing game design elements specifically.
Just as players and designers form models of the games they play, so must our systems.

Wardrip-Fruin argued that accounts of human interaction with games must concern both how games function procedurally \emph{and} how they communicate their operations to players~\cite{wardrip-fruin_playable_2005}.
We therefore proceed using his interpretive framework of \emph{operational logics}.
We see this as a key step forward for game design learning, as prior work has mainly considered games as bags of parameterized mechanics~\cite{zook_generating_2014,gregory2015grl}.
This approach can be effective, but it is not connected to player experience and, worse, it assumes mechanics as the fundamental building blocks whereas most mechanics are built out of several smaller pieces (e.g.\ collision detection, resource transactions, and other constraints and effects).

\subsection{Operational Logics} \label{sec:ol}

An operational logic combines an abstract process (e.g.\ agents that change behaviors over time, resolving collisions between embodied agents, and discrete transactions of numeric resources) with a strategy for communicating that process's behavior to players (respectively visual feedback for the current agent state, projections of bounding boxes onto a 2D plane, or animated readouts of values and flows)~\cite{mateas_defining_2009}.
Since these logics can be enumerated~\cite{osborn2017refining} and are often shared extensively across games, we can build AI that reifies these and operates simultaneously at the level of underlying abstract model \emph{and} what information is presented to users and how.

Games that use the same operational logics in similar ways (especially towards modeling the same systems) are more similar than they are different, even if the mechanics are quite distinct.
Consider the way that combat functions in different role-playing games: even if there are no specific rules or source code shared between games, we understand that characters' relative capabilities and status can be quantified in a compact schema; that combat can be resolved by turn-taking and the exchange of atomic resource transactions; and so on.
Such models transfer readily across games and even genres.
Players can quickly learn one game after playing another by adapting their pre-existing models along operational-logic lines.

We could use operational logics as a foundation for a knowledge representation: in fact, several successful projects in game generation have already done so.
On the modeling side, Martens aimed the Ceptre language at concretely specifying the abstract processes of OLs~\cite{martens_ceptre:_2015}; Game-o-Matic's underlying knowledge representation is also grounded in OLs~\cite{treanor2012game}.
Proceduralist readings (based on OLs) were recently operationalized in a hybrid generator/classifier~\cite{martens2016proceduralist}, generalizing Game-o-Matic's approach.
Our approach (enacted in the projects we discuss in Sec.~\ref{sec:learn_behaviors}) is to decide which logics we care to address and then select intermediate representations and learning techniques that support those logics.
Here we provide some examples; in most cases, algorithms already exist to learn the described structures, and there are clear player-side UI conventions or other affordances that communicate the concepts of interest.
\begin{description}[align=left]
\item [Physics Logics] --- Equations of motion can be learned by e.g.\ fitting quadratic curves to character positions in a segmented regression framework.
\item [Resource Logics] --- Model the movements of distinct types of resources between named locations using Petri nets or other numerical transition systems.
\item [Collision Logics] --- Identify game characters' bounding volumes and how characters impede each others' movement.
\item [Linking Logics] --- Organize a space (physical or metaphorical) as a graph of locations and directed edges, with ways to merge similar locations. 
\item [Character-State Logics] --- Learn state machines by positing one state per distinct animation or e.g.\ physics behavior, where transitions are guarded on conditions from other logics and with ways to merge similar states.
\item [Chance Logics] --- Probabilistic extensions of the above formalisms, potentially via probabilistic programming. 
\end{description}

\subsection{Observations}

The second question to be asked of any computational-intelligence system after ``How does it understand the world?'' must be ``How does it observe the world?''
In the most constrained case, we can only observe video of games being played; some of the works cited above have obtained good results from only this data source~\cite{guzdial_toward_2015,summerville2016learning}.
If we also have access to player inputs (e.g.\ a timed sequence of button presses), we can attempt causal reasoning, blaming changes in character behavior on player actions~\cite{summerville2017charda}.
Sometimes we can observe the game's internal runtime behavior including its memory address space or, for games run in emulation, the states of memory and registers of the emulated hardware~\cite{summerville2017mario}.

We might also control the platform on which the game runs, which allows us to manipulate memory values or drive new input sequences on-demand, saving and loading memory states to jump around and perform search.
There are also binary analysis tools like \(\mathrm{S}^2\mathrm{E}\)~\cite{chipounov2011s2e} and \emph{angr}~\cite{shoshitaishvili2016state}, which are designed to exercise programs' hard-to-reach behaviors.

The above is about as much as one can hope for given a black-box game program.
On the other hand, sometimes source code is available as well; in such cases we can deploy programming-language-specific model checkers and static analysis tools, or we might augment the approaches above with source-code level knowledge (e.g., blaming observed in-game interactions on specific lines of source code).
We are also interested in exploring the extent to which in-game text, game walkthroughs, or instruction manuals can bootstrap learning (as in~\cite{branavan2011non}), as can guessing the valence or semantics of visible game entities based on their appearance.
These are important information channels for human players and this kind of cultural or meta-game knowledge can be readily incorporated in the framework of OLs. 

\subsection{Human guidance}

Videogames by nature have extremely broad and deep search trees.
Fortunately, in many cases human players record their explorations through games as speedruns (often down to the level of timed input sequences).
Moreover, there exist archives of game save files and saved system states, admitting easy bookmarking of interesting regions of a game's possibility space even if the concrete path to get there is not recorded.

A key insight is that we can use these \emph{savestates} to achieve much better exploration of the possibility space than naive search alone could produce.
If we have a playthrough (either from a motivated player or from a designer), we could uniformly sample along prefixes of that playthrough (or randomly mutate it) to explore different parts of the game easily; this gives us a narrow global path from which we can branch out locally to investigate the nearby area.
In the design support context, we can use the game creators' previous playthroughs and manual testing as starting points for automated exploration, bootstrapping verification or rule-learning/updating tasks.

\section{Dynamic Analysis of NES Games}

We focus our work on the NES platform because fast software emulators are readily available, the hardware and many individual pieces of software are well understood, it has an extremely diverse catalog of games, and it is extremely well-known.
Generally, all our work proceeds by linking an emulator runtime, loading up a game program (colloquially, a \emph{Read-Only Memory (ROM)}), executing a series of inputs, and examining the emulated system's state by interrogating the emulator's runtime data structures.
%Many emulators support checkpoint save and restore operations, taking a snapshot of the whole system state that can later be restored at any time.
%This admits the use of search techniques as well as speculative execution along one or many possible futures.

While the NES has a general-purpose CPU and RAM, a key aspect that we take advantage of is its dedicated processor for image generation.
The Picture Processing Unit (PPU) has its own associated memory holding the data structures necessary for tiled rendering and sprite drawing.
This lets us extract ``visual information'' without doing extensive image processing or background subtraction.
We can also take advantage of information like color palettes, shared tile indices, and horizontal or vertical flipping to bias rule learning.  Furthermore, the separate treatment of tiled graphics and sprites helps us easily distinguish between game characters and backgrounds (this is complicated when the perceived characters are built mainly from background tiles, as in certain boss fights in \emph{Mega Man 2}).
The PPU has drawbacks: hardware sprites are always 8 pixels wide and either 8 or 16 pixels tall.
Most game characters are larger than this; Super Mario is 16 pixels wide and 24 pixels tall.
We therefore developed some sprite tracking algorithms which we detail elsewhere~\cite{summerville2017mario}.

\subsection{Learning Behaviors}
\label{sec:learn_behaviors}

Level layouts and character identities are structural properties which humans can often resolve just by looking at still images.
This work becomes more interesting when we consider learning dynamic design properties like the interactions between characters and the environment or character's dynamic behaviors.
\emph{glEnd() of Zelda} shows that simple heuristics suffice to understand many interactions with background tiles~\cite{murphy2016glend}; the rigor provided by the operational logics framework gives us more tools to distinguish types of tiles without enumerating in advance all the possible tile types.
\fxnote{Should we add something specific about plans/ways to handle identifying tile types/collisions/enemies vs non-enemies/etc?  Or identifying resource pickups?}

%\subsubsection{Character behaviors}

\begin{figure}[hbtp!]
        \centering
	        \includegraphics[width=0.49\textwidth]{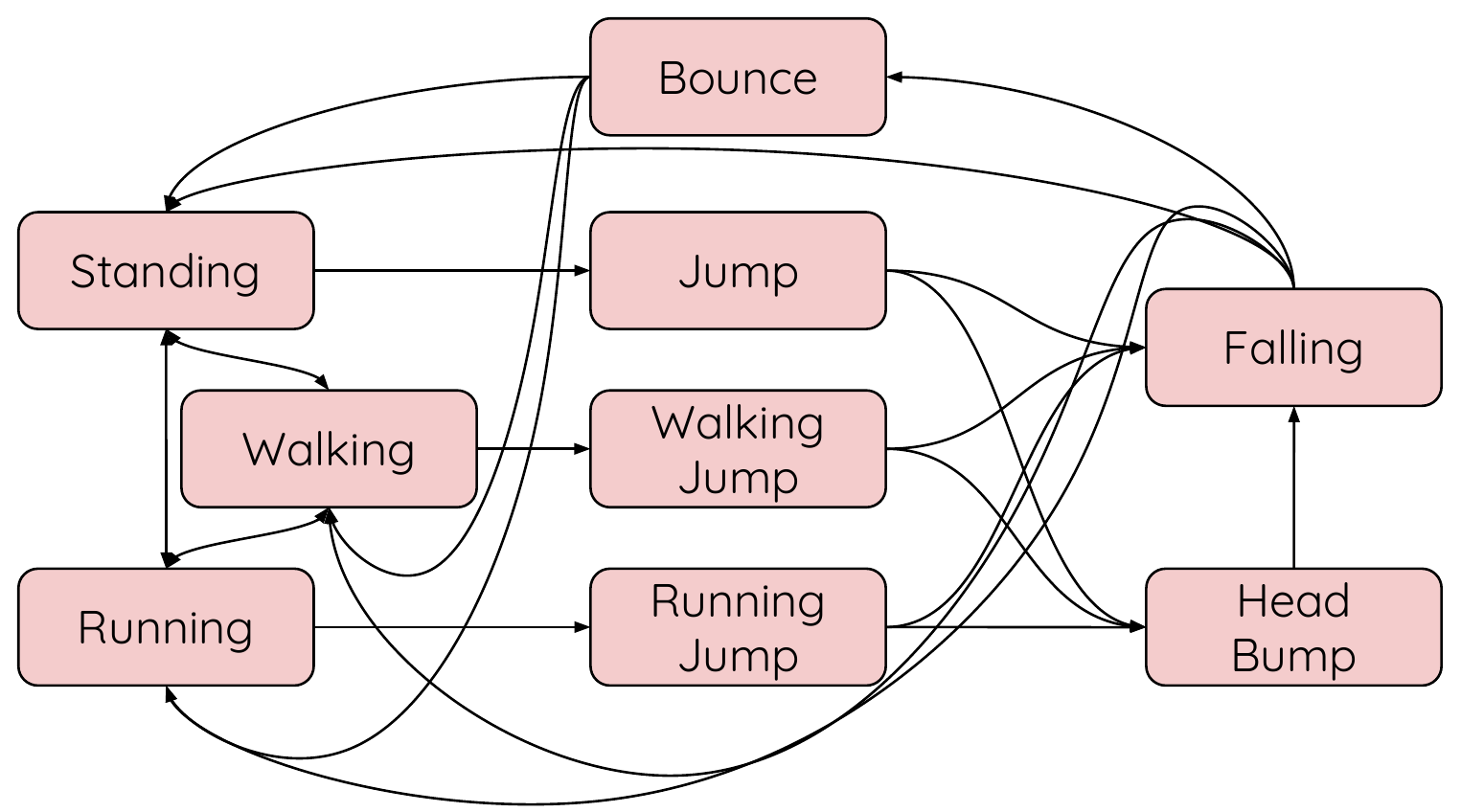}
        \caption{A finite state machine representing the movement rules governing Mario in \textit{Super Mario Bors.}}
        \label{fig:SMB_FSM}
\end{figure}
A specific composition of character-state, physics, and collision logics was the foundation for our earlier work in the CHARDA~\cite{summerville2017charda} and MARIO~\cite{summerville2017mario} projects.
Collision logics communicate to players that virtual objects may touch, and that events may transpire because of those collisions; we treat this as a causal learning problem where causes include the possible collisions between different types of objects and effects come from some other operational logic (but include the cessation of movement).
Physics logics describe the continuous movement of objects, which is naturally modeled with differential equations or their closed-form functions of time and initial state.
Character-state logics communicate that a character's behavior changes discretely between a finite set of states, and are captured well by finite state machines (e.g.\ see figure \ref{fig:SMB_FSM}); it is very likely that changes in a character's animation or physics behavior correspond to changes in character state and vice versa.

MARIO~\cite{summerville2017mario} assumes a per-character state machine structure and ignores collisions, learning only parameters on the physics equations associated with each state; CHARDA~\cite{summerville2017charda} generalizes this to learn state machine structure and physics parameters, with collisions being among the possible causes for transitions between discrete states (along with the axis and button inputs of input logics).
A reasonable extension for games like \emph{Mega Man} or \emph{Metroid} where some behaviors require and expend a resource like health or ammunition would be to incorporate resource transactions as a possible set of effects (and augment our causal language with resource availability); such resources can generally be treated abstractly as full, sufficient, or insufficient for particular outcomes, and these qualitative constraints should be straightforward to learn.

\section{Next Steps}

This vision suggests several new research areas that would have been tedious or infeasible without automation.
We also propose a few natural extensions for future work.

\subsection{Quantitative game studies}

The field of game studies relies on extensive play and examination~\cite{juul2007swap,treanor2011burgertime}.
This demands a large amount of a researcher's time and limits the scope of possible analyses, particularly constraining the number of games that can be investigated at once.
Fasterholdt \textit{et al.}~\cite{fasterholdt} studied a number of platformer games to form a general model of jumping, but their (labor-intensive) analysis was limited to four games.
Our work building on this~\cite{summerville2017mario} automatically performed in-game experiments, allowing us to capture and juxtapose the jumping dynamics of 48 games, a 12-fold increase.
Large data-based game studies~\cite{gamesage,xavier} generally rely on textual data such as Wikipedia or GameFAQs.
We believe that similar quantitative work extended into the actual dynamics of the games (not just text about the games) is an exciting new direction.

\subsection{Data-Driven Procedural Content Generation}

Data-driven Procedural Content Generation (PCG) has become a popular research field in recent years, but a key drawback is the lack of usable, clean data sources.
Projects such as the Video Game Level Corpus (VGLC) have begun to remedy this, but even this corpus only addresses 13 games from 7 different series.
While there are large databases of game levels as images~\cite{vgatlas} or videos (on YouTube), they lack much of the information required to accurately learn game mechanics, most notably control information.
Furthermore, while some automated processes of parsing these exist~\cite{videoparser}, they rely heavily on human input and annotation to be able to understand any of the mechanical properties of the level information.
Automatic gathering of level and behavior data from the game, combined with property-directed testing of assets for mechanical properties could allow for a much larger and  richer set of data for Data-driven PCG.  

Beyond levels, there is a broad variety of content that has yet to be generated in a data-driven manner including entities, mechanics, level progressions, and so on.
Due to the aforementioned reliance on computer vision and the lack of models of interaction, these have been impossible---until now.  

\subsection{Generalizing AGDL}

AGDL's current incarnation is tied to the NES and to found play traces.
Generalizing this to other platforms, including the Super NES, PuzzleScript, or even engines like Unity would be straightforward in some ways (the abstraction-learning code of CHARDA and MARIO would be largely unchanged), but it would also yield new challenges and opportunities.
Imbuing the design-learning agents with the ability to explore the game's possibility space on their own is also a natural next step: whether this means branching out from different instants of a given play trace or exploring from the beginning of the game, relational-learning agents, heuristic selection of experiments to try (like MARIO or \emph{glEnd() of Zelda}) and curiosity-driven search~\cite{gravina2016surprise,pathakICMl17curiosity} could help learn a broader variety of game rules more precisely.
Finally, addressing more types of operational logics is a clear step forward: resource logics, game-mode logics, and progression logics are both widespread and valuable to learn, as are the non-spatial linking logics used in game dialog trees.

\section{Conclusion}

In this paper we have described AGDL, a nascent field of research aiming to automatically learn useful, portable representations of game rules and instantial assets via observation and interaction with the game itself.
This is a productive research agenda that has been approached obliquely but rarely engaged directly.
AGDL has deep roots criss-crossing game design support, reverse engineering, statistical learning, inductive logic, specification recovery, and general game playing.
It presents a fascinating problem with obvious and significant applications; we hope that others will join us in exploring its boundaries and promise.

\bibliographystyle{IEEEtran}
\bibliography{IEEEabrv,agdl}

\end{document}